\documentclass[sigconf,balance]{acmart}

\usepackage{xcolor}
\usepackage{array}
\usepackage{booktabs}
\usepackage{multirow}

\AtBeginDocument{%
  }

\setcopyright{acmlicensed}
\copyrightyear{2018}
\acmYear{2018}
\acmDOI{XXXXXXX.XXXXXXX}
\acmConference[Conference acronym 'XX]{Make sure to enter the correct
  conference title from your rights confirmation email}{June 03--05,
  2018}{Woodstock, NY}
\acmISBN{978-1-4503-XXXX-X/2018/06}

\begin{document}

\title{Are Synthetic Videos Useful? A Benchmark for Retrieval-Centric Evaluation of Synthetic Videos}



\author{Zecheng Zhao}
\affiliation{%
  \institution{The University of Queensland}
  \city{Brisbane}
  \country{Australia}}
\email{uqzzha35@uq.edu.au}

\author{Selena Song}
\affiliation{%
  \institution{The University of Queensland}
  \city{Brisbane}
  \country{Australia}}
\email{selena.song@student.uq.edu.au}

\author{Tong Chen}
\affiliation{%
  \institution{The University of Queensland}
  \city{Brisbane}
  \country{Australia}}
\email{tong.chen@uq.edu.au}

\author{Zhi Chen}
\affiliation{%
  \institution{The University of Southern Queensland}
  \city{Toowoomba}
  \country{Australia}}
\email{zhi.chen@unisq.edu.au}

\author{Shazia Sadiq}
\affiliation{%
  \institution{The University of Queensland}
  \city{Brisbane}
  \country{Australia}}
\email{shazia@eecs.uq.edu.au}


\author{Yadan Luo}
\affiliation{%
  \institution{The University of Queensland}
  \city{Brisbane}
  \country{Australia}}
\email{y.luo@uq.edu.au}








\renewcommand{\shortauthors}{Zhao et al.}

\begin{abstract}
Text-to-video (T2V) synthesis has advanced rapidly, yet current evaluation metrics primarily capture visual quality and temporal consistency, offering limited insight into how synthetic videos perform in downstream tasks such as text-to-video retrieval (TVR). In this work, we introduce SynTVA, a new dataset and benchmark designed to evaluate the utility of synthetic videos for building retrieval models. Based on 800 diverse user queries derived from MSRVTT training split, we generate synthetic videos using state-of-the-art T2V models and annotate each video-text pair along four key semantic alignment dimensions: Object \& Scene, Action, Attribute, and Prompt Fidelity. Our evaluation framework correlates general video quality assessment (VQA) metrics with these alignment scores, and examines their predictive power for downstream TVR performance. To explore pathways of scaling up, we further develop an Auto-Evaluator to estimate alignment quality from existing metrics. Beyond benchmarking, our results show that SynTVA is a valuable asset for dataset augmentation, enabling the selection of high-utility synthetic samples that measurably improve TVR outcomes. Project page and dataset can be found at \url{https://jasoncodemaker.github.io/SynTVA/}.

\end{abstract}




\maketitle

\begin{figure}[!t]
    \centering
    \includegraphics[width=0.8\linewidth]{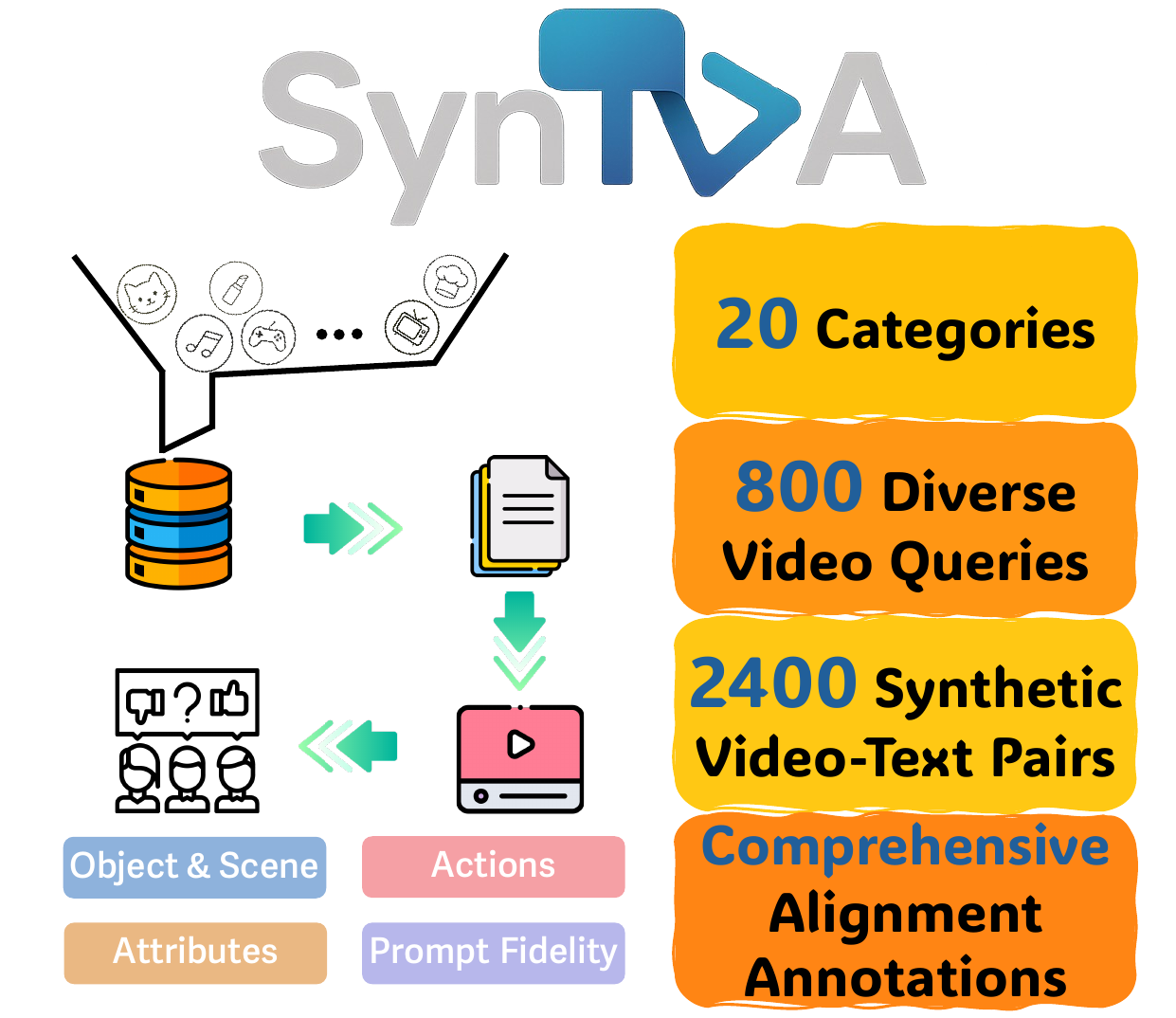}\vspace{-2ex}
    \caption{Overview of the proposed SynTVA dataset, designed to evaluate the usability of synthetic videos for downstream text-to-video retrieval tasks.}\vspace{-2ex}
    \label{fig:teaser}
    \Description{}
\end{figure}

\section{Introduction}

Text-to-video (T2V) synthesis has rapidly emerged as a frontier in generative modeling, enabling the creation of temporally coherent and semantically rich videos from natural language prompts. Recent advances are largely driven by diffusion-based architectures \cite{ho2020denoising_ddpm,peebles2023scalable_dit}, leading to a surge of competitive T2V models \cite{ho2022video_T2V, singer2022make_T2V, hong2022cogvideo_T2V, villegas2022phenaki_T2V, wan2025wan, agarwal2025cosmos, brooks2024video_sora, mochi} from both industry and academia: Sora \cite{brooks2024video_sora} exemplifies a \textit{model-centric} approach by leveraging diffusion transformers (DiT) to synthesise high-resolution and long-duration videos with strong prompt fidelity; Mochi \cite{mochi} introduces an asymmetric DiT architecture that decouples spatial and temporal modeling to better capture motion dynamics over time. In contrast, Cosmos \cite{agarwal2025cosmos} takes a data-centric strategy by constructing a world foundation model trained on large-scale, physics-aware video corpora for broader generalisation; Wan2.1 \cite{wan2025wan} proposes a 3D causal VAE that explicitly models temporal dependencies through latent dynamics, improving motion consistency and temporal reasoning.

The boom in T2V models has created an urgent need for a comprehensive evaluation framework. Traditional metrics such as Fréchet Inception Distance (FID) \cite{FID} and Fréchet Video Distance (FVD) \cite{FVD1, FVD2}, while widely used, are often misaligned with human perception and fail to capture the temporal and semantic complexities of videos. Similarly, conventional video quality assessment (VQA) methods \cite{VQA, VQA2} tend to focus on low-level distortions and lack the granularity required for generative tasks. In response, recent benchmarks such as EvalCrafter, VBench, and T2VBench \cite{EvalCrafter, VBench, T2VBench} have proposed more holistic evaluation protocols. For instance, EvalCrafter \cite{EvalCrafter} introduces a multi-dimensional framework that assesses visual quality, motion amplitude, temporal consistency, and text-video alignment. VBench \cite{VBench} further extends the evaluation space with 16 fine-grained dimensions and a diverse prompt set that emphasises prompt adherence and perceptual realism.

However, despite these advancements, a critical gap remains: \textit{Do high scores on these evaluation benchmarks reflect actual utility in downstream tasks?} Current evaluations primarily emphasise perceptual or stylistic aspects, reflecting how humans perceive the generated content. Yet, a key application of synthetic videos lies in data augmentation for machine learning models (eg., autonomous driving \cite{richter2016playing_drive, dosovitskiy2017carla_drive} and action recognition \cite{varol2017learning_action, petrovich2022temos_action}). In this context, their true value is interpreted through improvements in model performance rather than subjective human assessment. This potential disconnect means that the relationship between high \textit{visual quality} (according to current metrics) and \textit{high utility} (for model training) remains unclear. Such uncertainty casts doubt on the ability of existing metrics to effectively guide the selection and generation of synthetic data for practical, real-world applications. To probe this utility, text-to-video retrieval (TVR) \cite{jin2024mv_tvr, gorti2022x_tvr, liu2022ts2_tvr, xue2022clip_tvr, deng2023prompt_tvr, luo2022clip4clip_tvr, li2023progressive_tvr, jin2023diffusionret_tvr} offers a pertinent application, as it directly relies on robust cross-modal alignment, a principle also fundamental to T2V generation.

To address this challenge and form an answer, we introduce SynTVA (\textbf{Syn}thetic video-text dataset featuring strong \textbf{T}ext-\textbf{V}ideo \textbf{A}lignment), a new benchmarking dataset specifically designed to assess the utility of synthetic videos in TVR tasks. Our benchmark moves beyond traditional appearance-focused metrics to evaluate whether synthetic videos can meaningfully support retrieval tasks in realistic query settings. As illustrated in Figure~\ref{fig:teaser}, we begin by leveraging GPT-4o \cite{gpt4o} to generate 800 user queries across 20 semantic categories. These queries are guided by the MSRVTT training split \cite{MSRVTT}, a widely used TVR benchmark. The queries serve two purposes: (1) align with MSRVTT’s real-world search intents for practical relevance, and (2) extend into underrepresented concepts to improve TVR via semantic augmentation.


To generate the corresponding videos, we synthesize video prompts using state-of-the-art open-source T2V models, including Cosmos \cite{agarwal2025cosmos}, Mochi \cite{mochi}, and Wan2.1 \cite{wan2025wan}. This synthesis yields a total of 2,400 videos, which collectively amount to 200 minutes of visual content. Each video-text pair is then evaluated by five human judges across four core criteria critical for retrieval: \textit{Object \& Scene}, \textit{Actions}, \textit{Attributes}, and \textit{Prompt Fidelity}. The entire annotation effort exceeded 100 hours. These dimensions are selected to reflect key semantic aspects that users typically rely on when searching for video content, enabling a more task-relevant evaluation of T2V utility.

To systematically study the relationship between appearance-focused VQA metrics and functional alignment for TVR, we introduce a unified evaluation framework. This framework first computes a range of standard VQA metrics and then analyzes their correlation with our retrieval-centric alignment scores. To further bridge the gap between these VQA metrics and actual TVR utility, we develop an Auto-Evaluator. This tool models the relationship between appearance quality and retrieval-relevant semantic alignment by learning to predict human alignment scores for the critical \textit{Object \& Scene} and \textit{Actions} dimensions, using a subset of features derived from VBench \cite{VBench}. This approach allows us to assess the predictive capacity of existing metrics for estimating this crucial task-relevant semantic alignment.

Finally, we conduct TVR experiments using high- and low-quality synthetic subsets ranked by human annotation scores to assess how the quality of semantic alignment in synthetic videos impacts their utility in downstream tasks. Extensive results reveal two key findings: (1) Videos rated highly effectively enhance TVR performance as training quantity increases. This suggests that, through our Auto-Evaluator, appearance-focused VQA metrics from VBench \cite{VBench} are demonstrated to have a positive correlation with TVR utility. (2)~Among the four alignment dimensions, \textit{Object \& Scene} and \textit{Actions} exhibit the most substantial impact on TVR outcomes, identifying them as critical factors for synthetic video utility in retrieval tasks. These insights offer valuable guidance for future research, particularly in methods for selecting or generating high-utility synthetic video samples, and for developing more effective T2V models geared towards robust downstream applications.

\begin{figure*}[!tp]\vspace{-1ex}
    \centering
    \includegraphics[width=1\linewidth]{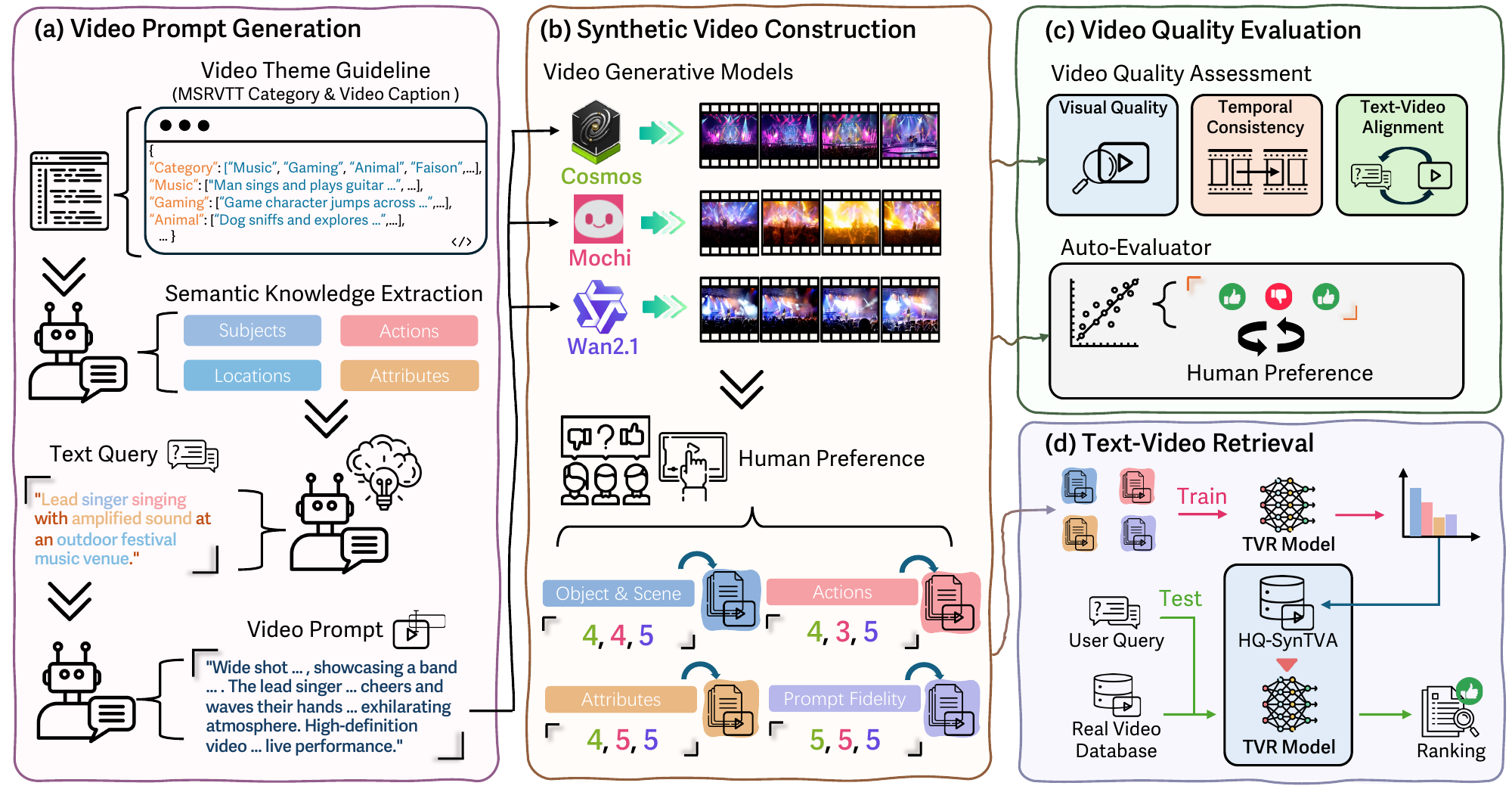}\vspace{-2ex}
    \caption{Overview of the SynTVA pipeline and its application.\textmd{ (a) Video Prompt Generation: GPT-4o \cite{gpt4o} generates diverse prompts from MSRVTT-based themes by extracting and expanding semantic elements. (b) Synthetic Video Construction: State-of-the-art T2V models (Cosmos \cite{agarwal2025cosmos}, Mochi \cite{mochi}, Wan2.1 \cite{wan2025wan}) generate videos from shared prompts, followed by human annotation across four alignment dimensions. (c) Video Quality Evaluation: Standard VQA metrics are used to assess quality, and an Auto-Evaluator predicts alignment with human preferences. (d) Text-to-Video Retrieval: SynTVA subsets with varied alignment qualities are used to train and evaluate TVR models. The curated HQ-SynTVA set improves retrieval performance using only synthetic data.}}
    \label{fig:workflow}
    \Description{}
\end{figure*}

\section{Dataset Overview}



We introduce SynTVA, a dataset of synthetic video-text pairs highlighting the robust semantic correspondence between the two modalities. To move beyond typical perceptual VQA metrics, which overlook practical utility, SynTVA is constructed as an emulation of the widely adopted MSRVTT dataset \cite{MSRVTT}. Our purpose is to investigate the impact of utilizing synthetic data for downstream TVR tasks.

As illustrated in Figure \ref{fig:workflow} (a) and (b), the dataset construction consists of the following two steps: (1) We utilize GPT-4o \cite{gpt4o} to generate detailed and varied user queries and video prompts that have the potential to generate candidate videos. Specifically, we extract and analyse the category labels and video captions in MSRVTT \cite{MSRVTT}. This ensures broad scenario coverage and promotes effective text-video alignment. (2) We employ three T2V models to transform the generated textual prompts into corresponding visual narratives. To evaluate semantic alignment between video and corresponding queries, we task human judges to annotate those video-text pairs based on four different perspectives.

\subsection{Video Prompt Generation}
To allow the video prompts with the potential to generate specific and narrow videos, we design a pipeline with three stages: 
(1) The 20 category labels in the MSRVTT training split \cite{MSRVTT} reflect common YouTube content, e.g., Music, Gaming, Animal, etc. For each category, given the category name and the associated video captions, we prompt GPT-4o to extract a vocabulary of candidate \textit{Subjects}, \textit{Actions}, \textit{Locations}, and \textit{Attributes}. 
(2) Guided by concise instructions, GPT-4o \cite{gpt4o} then recombines these elements to draft 40 novel scene queries per category. This results in a total of 800 distinct text queries.
(3) To turn each query into a full video prompt, we ask GPT-4o \cite{gpt4o} to append shot-level descriptors, including camera motion, atmosphere, stylization, and other cinematic cues. These enrichments give generative models stronger conditioning signals and enable fairer comparisons during video-quality assessment.





\vspace{-1ex}
\subsection{Synthetic Video Construction}

Given the generated video prompts, we synthesise videos using three state-of-the-art (SOTA) text-to-video generative models, i.e., Cosmos \cite{agarwal2025cosmos}, Mochi \cite{mochi} and Wan2.1 \cite{wan2025wan}. Each model is used to generate a five-second video at a minimum resolution of 480p for each one of the 800 prompts, resulting in a total of 2,400 videos. To the best of our knowledge, comprehensive video quality benchmarks for these three generative models are still lacking. While recent efforts have focused on visual fidelity or general plausibility, few have addressed semantic alignment from a human-centric perspective. Our dataset specifically targets text–video alignment quality, with a focus on practical applications in text-based video retrieval. Following the generation of the raw dataset, we enrich each text–video pair with fine-grained human annotations that assess how accurately the visual content reflects the semantics of the input text.




We design a comprehensive evaluation rubric to assess text-video alignment across four key aspects, each targeting a different dimension of semantic fidelity:
\begin{itemize}
    \item \textbf{Object \& Scene:} \textit{How well does the video depict the main objects and the overall setting described in the text?}
    \item \textbf{Action:} \textit{How well does the video depict the actions or movements described in the text?}
    \item \textbf{Attributes:} \textit{How well does the video depict the descriptive qualities (e.g., adjectives, adverbs) from the text?}
    \item \textbf{Prompt Fidelity:} \textit{Does the video introduce any extraneous elements that were not mentioned in the text query?}
\end{itemize}
Each of these dimensions is rated on a 5-point scale, where 1 signifies a `bad match' and 5 represents a `perfect match'. These four aspects effectively cover most user interests in text-based video searching and also consider the potential noise that could be introduced during TVR model training.


For each text query, we assign five human judges to evaluate all three synthetic videos generated by the different models. 
Each judge independently assesses the alignment between the text and each video across the four rubric dimensions described above. This results in fifteen annotated video evaluations per prompt, capturing diverse perspectives on semantic alignment.


In total, the SynTVA dataset comprises 800 diverse text queries, a total of 200 minutes of video, and 2,400 video–text pairs, with each pair accompanied by five sets of detailed human annotations across four alignment dimensions. SynTVA thus offers a high-quality benchmark for assessing aspects of semantic alignment that are critical for their utility in downstream retrieval tasks.

\vspace{-2ex}
\subsection{Dataset Analysis}

\begin{figure}[!t]
    \centering
    \includegraphics[width=1\linewidth]{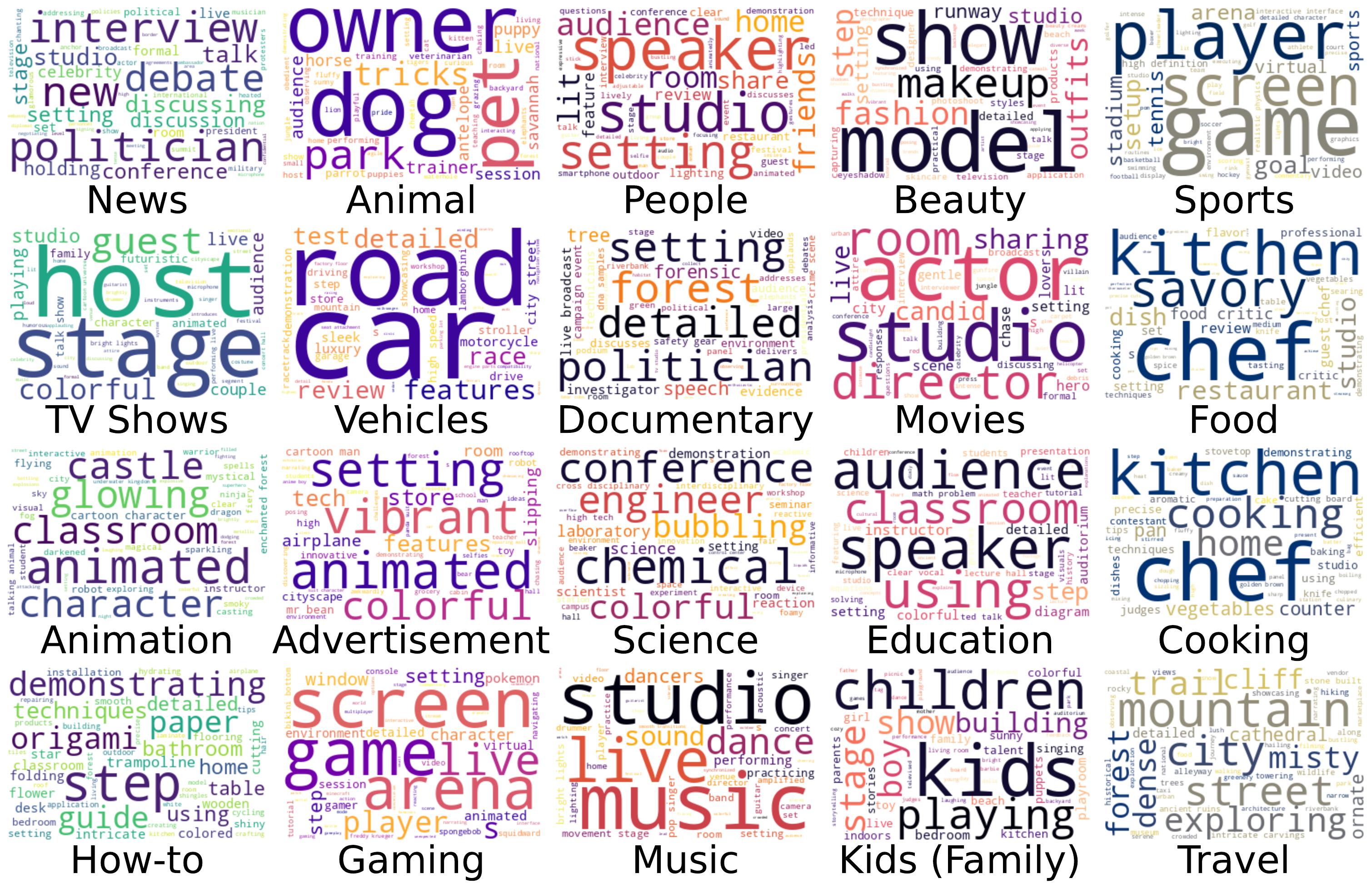}\vspace{-2ex}
    \caption{Per-category word clouds showcasing keywords within each of SynTVA's 20 diverse text query themes.}
    \label{fig:word-cloud}
    \Description{}
\end{figure}

\begin{figure}[!t]\vspace{-3ex}
    \centering
    \includegraphics[width=1\linewidth]{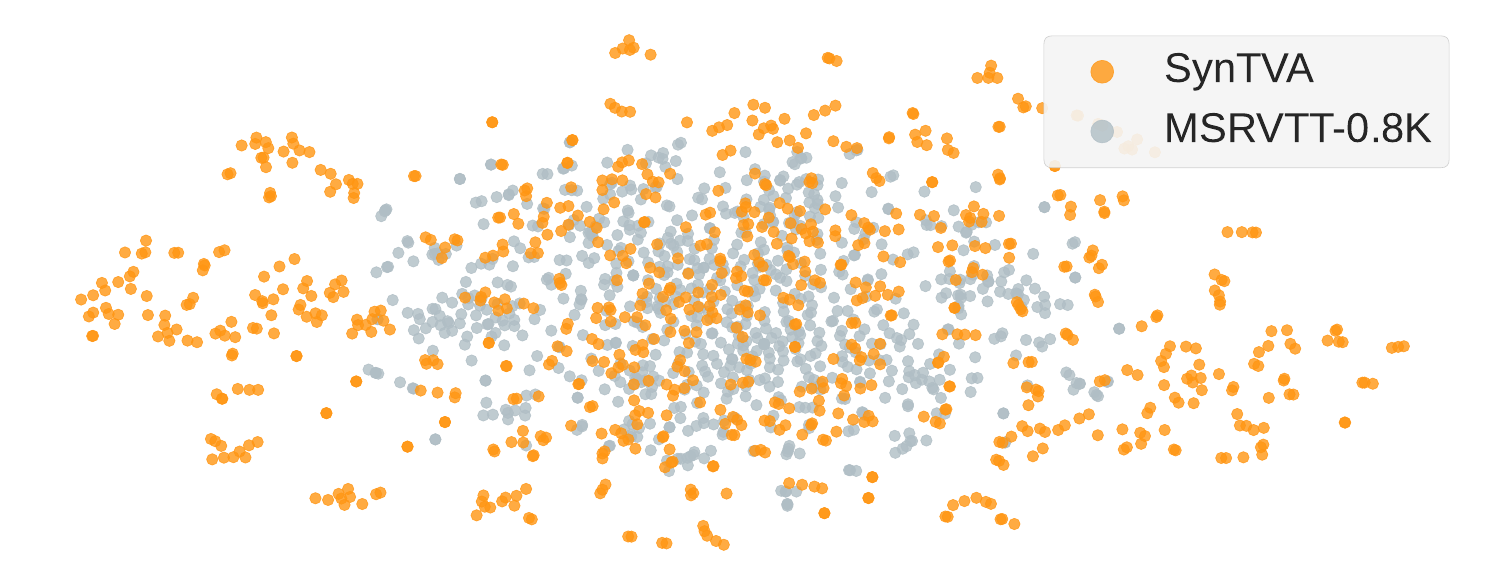}\vspace{-3ex}
    \caption{T-SNE visualisation comparing the text embedding distributions of our SynTVA and a randomly selected MSRVTT-0.8K subset.}
    \label{fig:embeddings_comparison}
    \Description{}
\end{figure}

\textbf{Diversity of Text Queries.} Our dataset has 800 unique text queries, covering 20 categories. As shown in Figure \ref{fig:word-cloud}, we illustrate the key terms and thematic focus within each category. The word clouds demonstrate thematic coherence within categories, while also reflecting a rich variety of distinct concepts. Moreover, Figure \ref{fig:embeddings_comparison} compares the distribution of text embeddings between our text queries and a randomly sampled MSRVTT-0.8K subset through T-SNE visualisation \cite{van2008visualizing}. It is clear that our text query achieves significantly wider semantic coverage than the MSRVTT-0.8K subset.

\begin{figure}[h]
    \centering
    \includegraphics[width=1\linewidth]{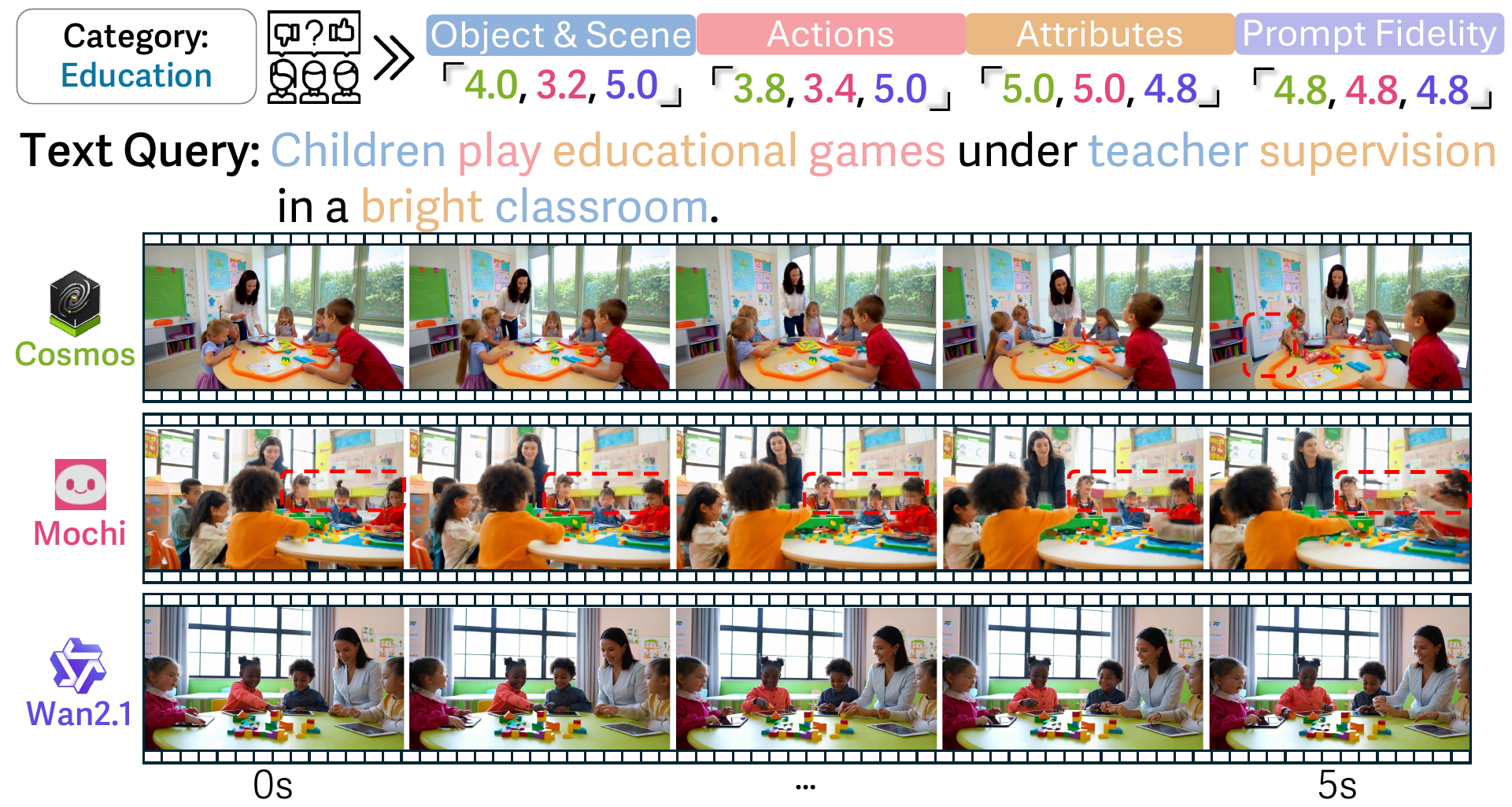}\vspace{-2ex}
    \caption{Case Study of T2V Model Outputs. \textmd{Comparison of Cosmos \cite{agarwal2025cosmos}, Mochi \cite{mochi}, and Wan2.1 \cite{wan2025wan} video generations for the same text query. }}
    \label{fig:case-study}
    \Description{}
    \vspace{-10pt}
\end{figure}

\noindent 
\textbf{Qualitative Study on T2V Model Outputs.}
We synthesise videos from three models for each text query. As shown in Figure \ref{fig:case-study}, all generated videos show reasonable semantic content corresponding to the given text query. However, in the video generated by Cosmos \cite{agarwal2025cosmos}, the girl to the left of the desk suddenly disappears in the last few frames. Additionally, in the video generated by Mochi \cite{mochi}, human faces are blurry, and its slow-motion makes the `playing game' action difficult to discern. In contrast, the video from Wan2.1 \cite{wan2025wan} is close to realistic footage. The human preference score aligns with the realism analysis of these videos.

\begin{figure}[h]
    \centering
    \includegraphics[width=1\linewidth]{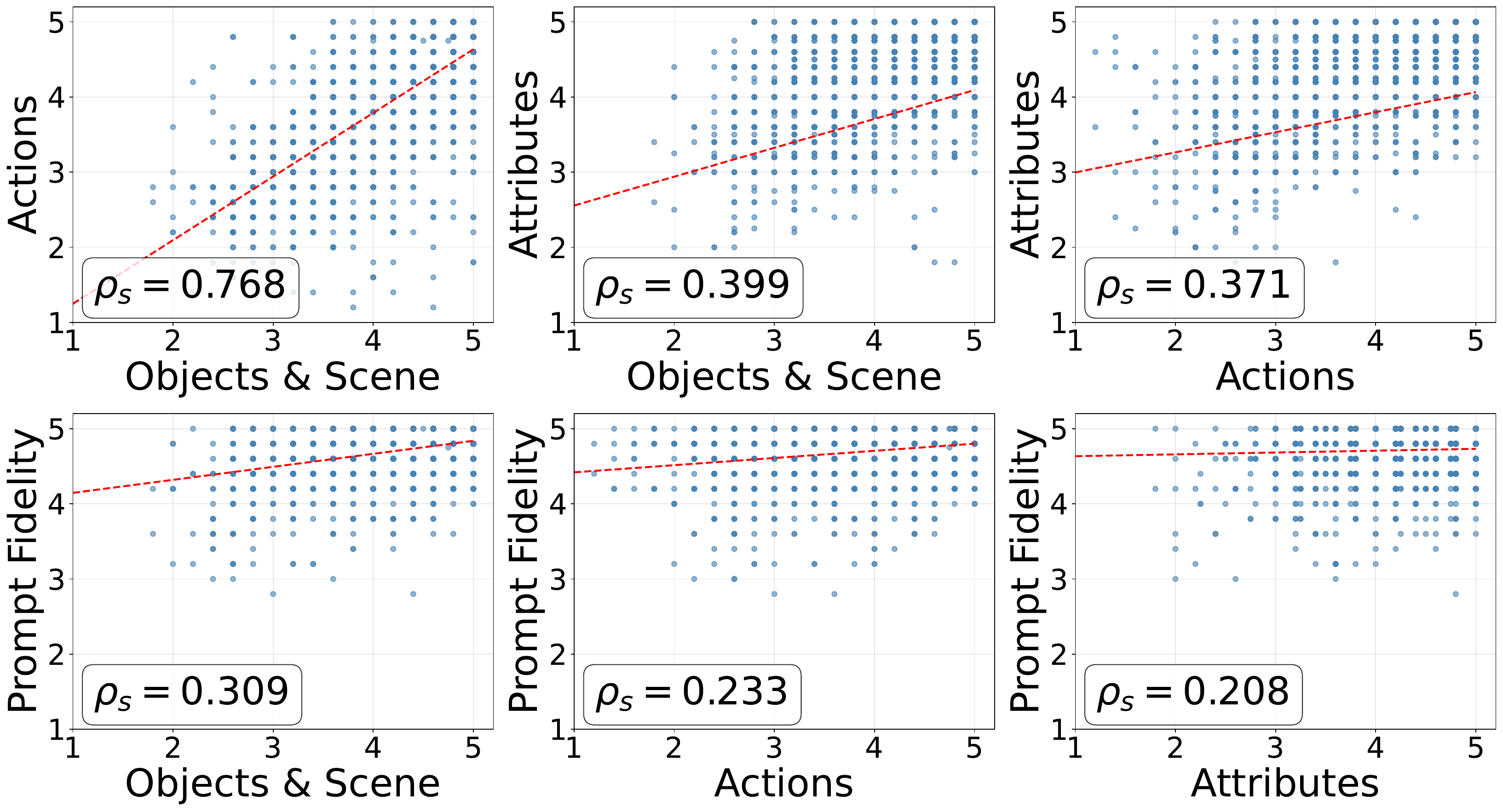}\vspace{-2ex}
    \caption{Correlation between the four evaluation aspects. \textmd{$\rho_s$ represents the spearman rank correlation coefficient}.}
    \label{fig:aspect_correlation}
    \Description{}
\end{figure}

\begin{figure}[!t]\vspace{-2ex}
    \centering
    \includegraphics[width=1\linewidth]{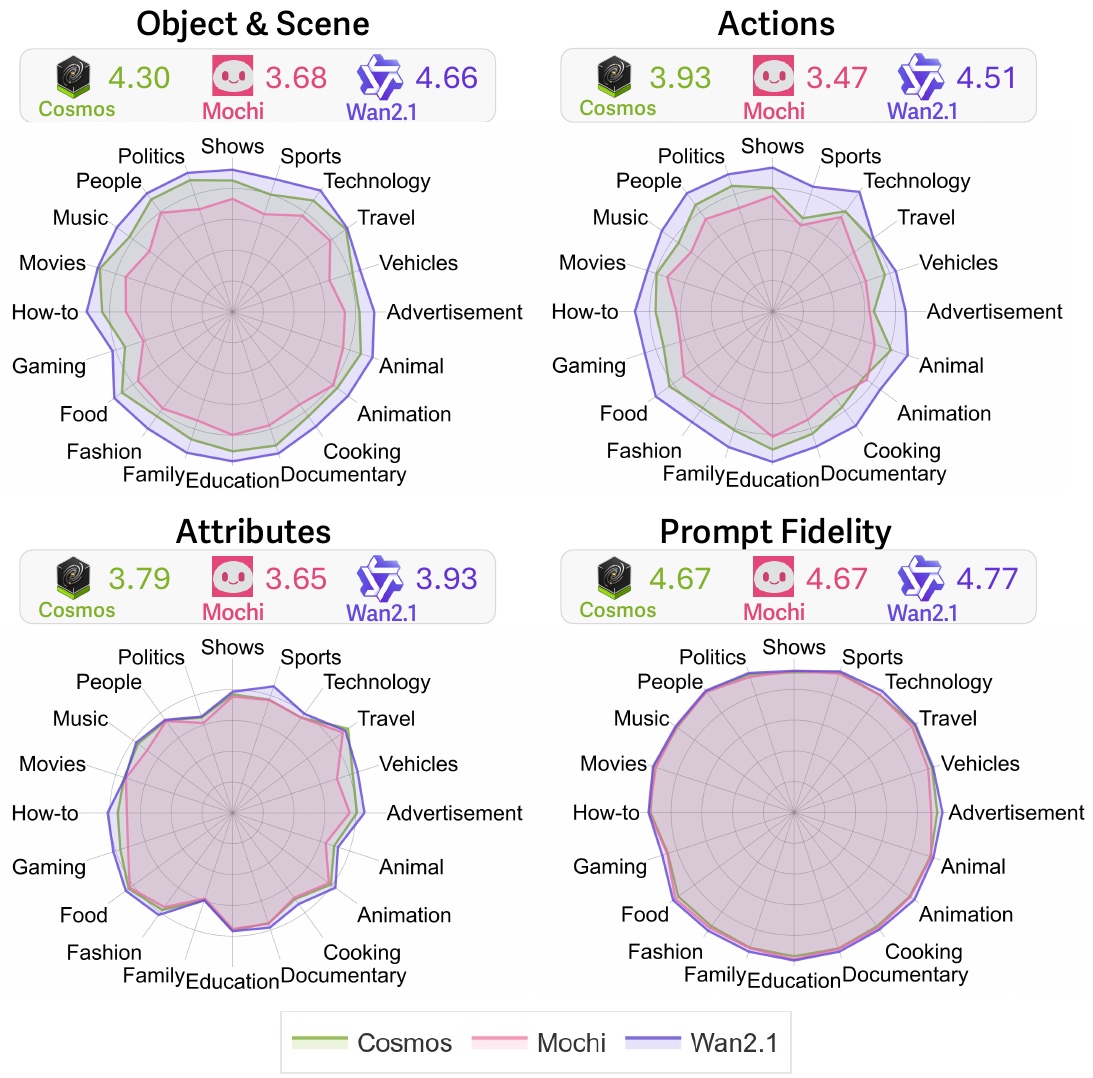}\vspace{-2ex}
    \caption{Comparison of the performance of T2V models across 20 categories for each text-video alignment aspect.}
    \label{fig:rader_chart}
    \Description{}
    \vspace{-10pt}
\end{figure}

\noindent 
\textbf{Analytics on Human Preference Annotation.} 
After cleaning and aggregating the human score annotations, we analyse the model performance from each aspect, as presented in Figure \ref{fig:rader_chart}. Wan2.1 \cite{wan2025wan} consistently outperforms the other models across all four aspects, while Mochi \cite{mochi} generally receives the lowest scores. Notably, all models achieve relatively high scores in the \textit{Object \& Scene} and \textit{Prompt Fidelity} aspects, indicating that SOTA video generative models can effectively adhere to video prompt guidance and accurately capture diverse object and scene semantics and their spatial relationship. However, the performance in the \textit{Attributes} aspect is generally lower. Through our observation, models exhibit a deficient semantic understanding of abstract descriptive qualities, particularly concerning how to effectively translate concepts like \textit{atmosphere} into visual representations. We also investigate the correlations among our four evaluation aspects, as shown in Figure \ref{fig:aspect_correlation}. A strong positive Spearman correlation ($\rho =0.768$) between the \textit{Object \& Scene} and \textit{Action}  suggests that when videos struggle to clearly depict primary objects, they are often similarly challenged in accurately representing the associated actions. In contrast, the correlations among other pairs of aspects are considerably weaker (all $\rho <0.4$), indicating these act as independent dimensions for evaluating distinct facets of text-video alignment.

As shown in Figure \ref{fig:rader_chart}, we also investigate the per-category performance across models. In the \textit{Object \& Scene} aspect, all models perform poorly in the \textit{Gaming} category. Analysis of human judges' comments reveals that these models lack the capability to generate specific gaming characters or intricate game scenes. Regarding \textit{Actions}, complex sports techniques and dynamic player interactions make it difficult for the models to generate precise action sequences. In the \textit{Attributes} aspect, queries for categories such as \textit{Family} and \textit{Politics} often include abstract descriptions of atmosphere, which demonstrate models' limited semantic understanding of abstract descriptive adjectives, such as \textit{warm} or \textit{passionate}.



\section{Experiments}
To close the critical gap concerning whether high scores on current evaluation benchmarks reflect actual utility in downstream tasks, we introduce a unified evaluation and experiments framework. As presented in Figure \ref{fig:workflow} (c) and (d), the framework aims to: (1) Probe the relationship between traditional appearance-focused VQA metrics and our retrieval-centric semantic alignment dimensions. (2) Develop an Auto-Evaluator that learns to predict human alignment scores for critical retrieval dimensions using existing VQA metrics as input. (3) Evaluate the real-world utility of synthetic videos by constructing SynTVA subsets with varying quality and quantity to train TVR models and assess their performance.

\vspace{-5pt}
\subsection{Evaluation}

\subsubsection{Video Quality Assessment}
\label{VQA}
To assess the utility of established VQA metrics within our alignment-focused framework, we investigate their relationship with our human-annotated scores. Therefore, we select the following three widely adopted VQA metrics for video evaluation:
\begin{itemize}
    \item \textbf{Visual Quality (Inception Score (IS) \cite{IS}):} It assesses the diversity of generated videos using a pre-trained Inception Network. A higher IS generally indicates a richer variety in visual output, and thus better visual quality.   
    \item \textbf{Temporal Consistency (CLIP-Score \cite{radford2021CLIP}):} We select CLIP-Score between frames to evaluate the semantic stability between consecutive frames in a video. It achieves this by calculating the average cosine similarity of CLIP embeddings from adjacent frames, and higher is better.
    \item \textbf{Text-Video Alignment (SD Score \cite{EvalCrafter}):} SD Score measures the text-video alignment by comparing generated video frames against images synthesized by a text-to-image model (SDXL \cite{podell2023sdxl}) from the text query, using CLIP embedding similarity. The approach aims to reflect content consistency between video and text.
\end{itemize}

\begin{figure}[!t]\vspace{-2ex}
    \centering
    \includegraphics[width=1\linewidth]{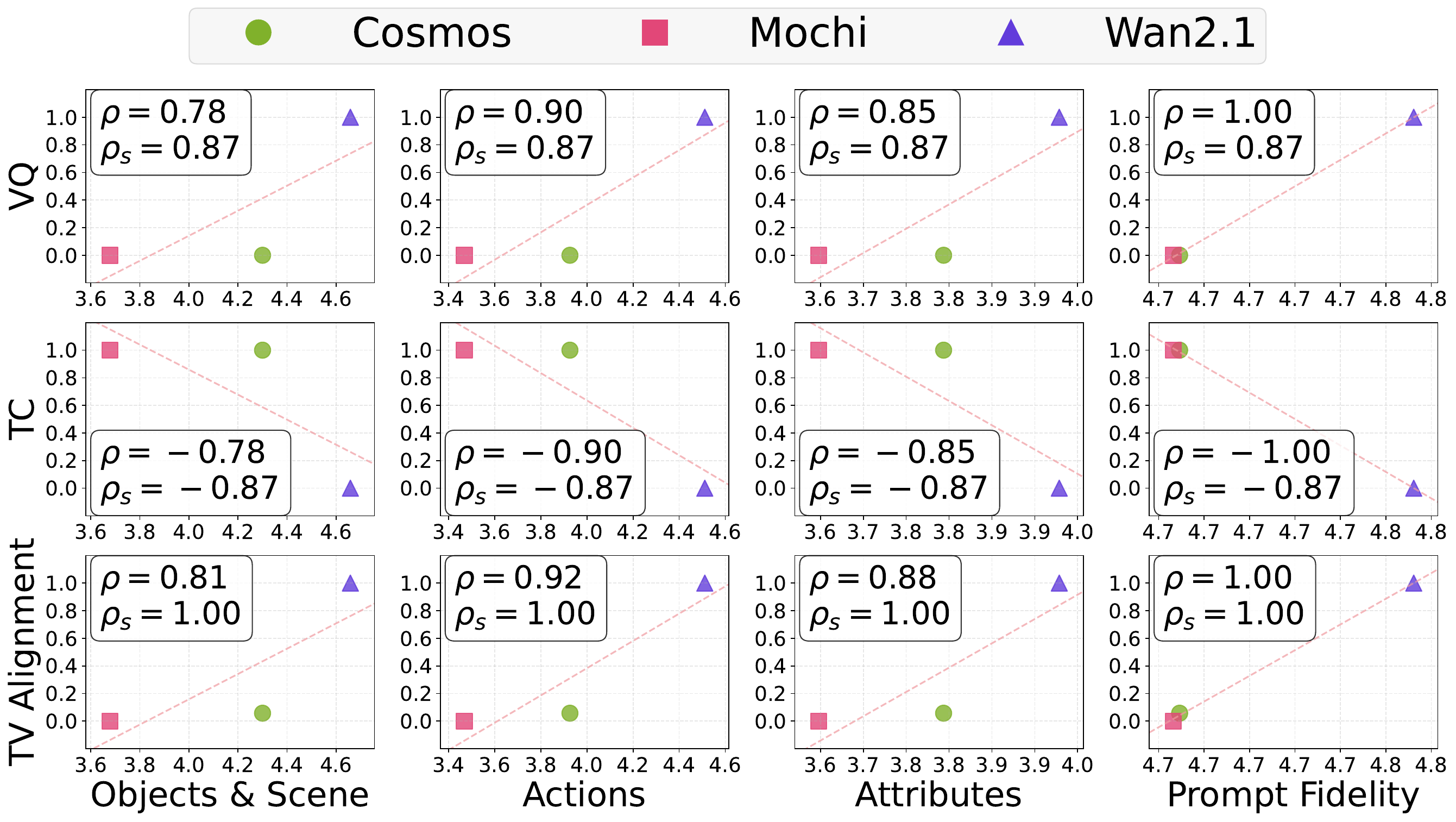}\vspace{-2ex}
    \caption{Correlations between VQA metrics and human alignment scores for different T2V models. \textmd{VQ: Visual Quality; TC: Temporal Consistency; TV Alignment: Text-Video Alignment}}
    \label{fig:correlationMatrix}
    \Description{}
\end{figure}

\noindent As shown in Figure \ref{fig:correlationMatrix}, the VQA metrics exhibit distinct relationships with our human annotations. \textit{Text-Video Alignment} demonstrates a strong and consistent positive correlation with all four aspects, supporting the validity of our metric design. \textit{Visual Quality}  also correlates positively, suggesting that the perceived alignment is influenced by the video clarity. In contrast, \textit{Temporal Consistency} shows negative correlations, indicating that temporal inconsistency may have a limited impact on human judgment of alignment.

\begin{figure}\vspace{-3ex}
    \centering
    \includegraphics[width=1\linewidth]{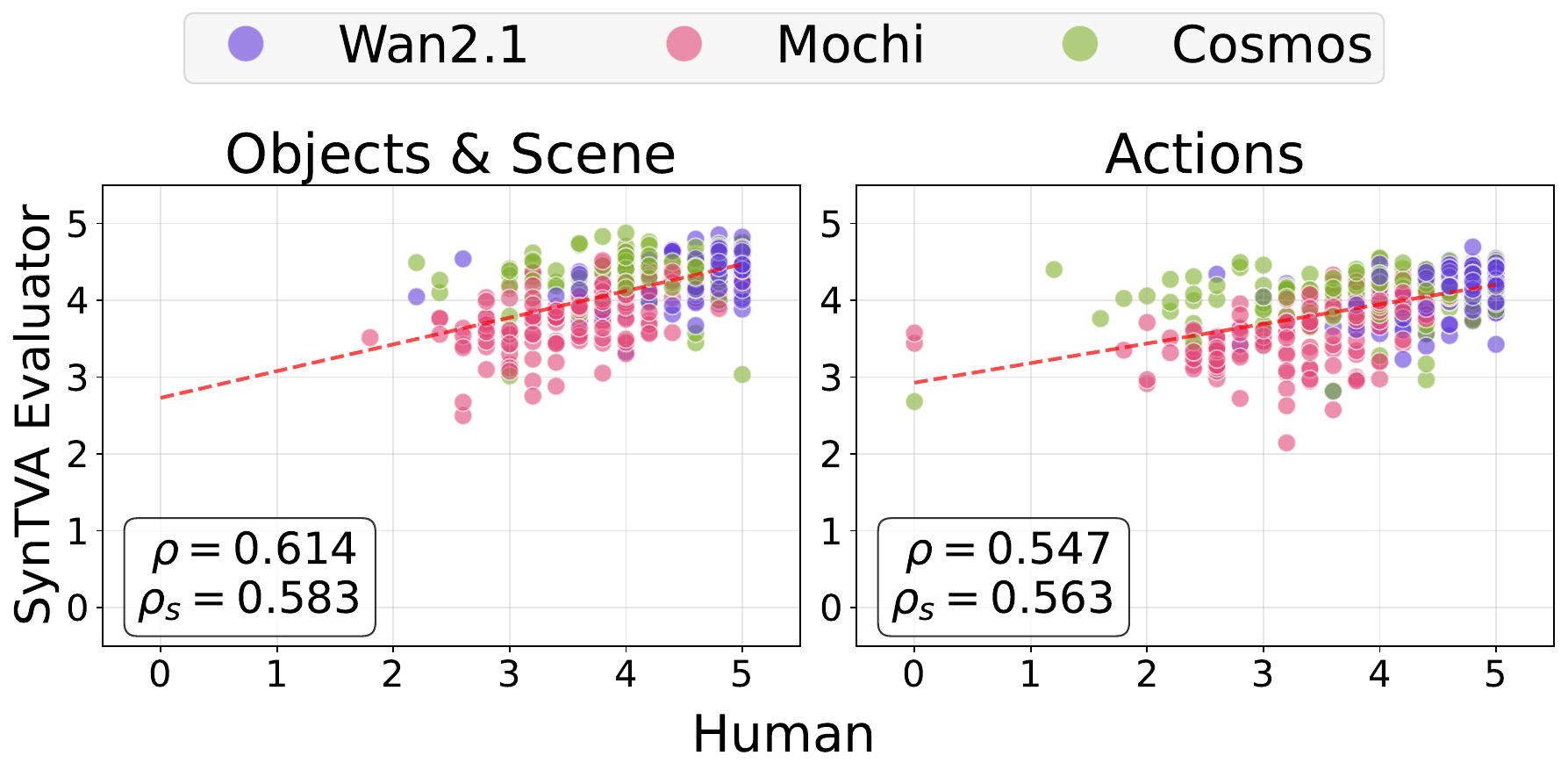}\vspace{-3ex}
    \caption{Performance of the SynTVA Auto-Evaluators. \textmd{The results show a strong positive correlation with human judgments on the key  dimensions of Object \& Scene and Action.}}
    \label{fig:auto_evaluator}
    \Description{}
    \vspace{-10pt}
\end{figure}

\begin{figure*}[!tp]
    \centering
    \includegraphics[width=1\linewidth]{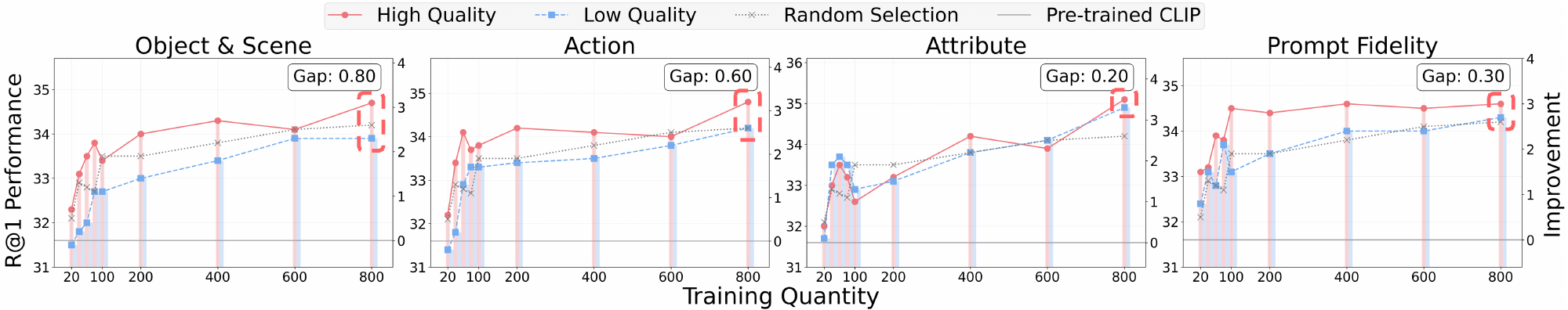}\vspace{-3ex}
    \caption{Impact of synthetic data quality and quantity on TVR R@1 performance across our alignment aspects.}\vspace{-2ex}
    \label{fig:TrainingQuantity}
    \Description{}
\end{figure*}

\subsubsection{Auto-Evaluator}


To efficiently assess synthetic video utility for TVR without costly human annotation, we develop Auto-Evaluator to predict human alignment scores from VQA metrics. Given that our TVR experiments (Figure \ref{fig:TrainingQuantity}) highlight the semantic alignment of \textit{Object \& Scene} and \textit{Actions} as critically impacting retrieval outcomes, we develop two specialized Auto-Evaluators to predict human scores specifically for these high-impact dimensions. 



To construct the two Auto-Evaluators, we train two linear regression models with seven general VQA metrics \cite{VBench} as input, and human scores on \textit{Object \& Scene} and \textit{Actions} as output, respectively. Specifically, the seven general VQA metrics include \textit{Image Quality} and \textit{Subject Consistency} for evaluating object-level quality; \textit{Background Consistency} for evaluating background fidelity; \textit{Aesthetic Quality}, \textit{Motion Smoothness} and \textit{Temporal Flickering} assess video quality from a global perspective; and \textit{Overall Consistency} represents text-video alignment. We utilize the 2,400 video-text pairs from the SynTVA dataset, which is split into 80\%/20\% for training and testing. The performance of these Auto-Evaluators is then evaluated by Pearson Correlation Coefficient ($\rho$) and the Spearman Rank Correlation Coefficient ($\rho_s$).

As shown in Figure \ref{fig:auto_evaluator}, Auto-Evaluators produce predictions consistent with human annotation scores. The correlation metrics $\rho$ and $\rho_s$ indicate the effectiveness in modelling human preferences via the selected VQA dimensions. Table  \ref{tab:vqa_metrics_analysis} presents the contribution weights of each VQA metric. For \textit{Object \& Scene}, quality-related metrics (\textit{e.g.,} Imaging, Aesthetic) and Consistency (\textit{e.g.,} Subject, Background) contribute positively, suggesting that humans judge the alignment based on the clarity and consistency across frames. Mirroring Section \ref{VQA}, \textit{Temporal Flickering} shows 0.00\% contribution, indicating it does not impact human recognition. For \textit{Actions}, \textit{Aesthetic Quality} and \textit{Subject Consistency} are most influential, implying that higher-quality visuals aid action interpretation. 
Meanwhile, \textit{Motion Smoothness} and \textit{Overall text-video Consistency} contribute 0.00\%, suggesting these are less critical for aligning actions with textual descriptions in general text-video alignment metrics.

\definecolor{lowsatred}{rgb}{0.8, 0.5, 0.5}
\begin{table}[!tp]
\caption{Contribution of individual VQA metrics to Auto-Evaluator performance across alignment dimensions.}\vspace{-2ex}
\centering
\renewcommand{\arraystretch}{0.9}
\scalebox{0.96}{\begin{tabular}{lcccc}
\toprule
\multirow{2}{*}{\textbf{VQA Metrics}} & \multicolumn{2}{c}{\textbf{Object \& Scene}} & \multicolumn{2}{c}{\textbf{Actions}} \\
\cmidrule(lr){2-3} \cmidrule(lr){4-5}
 & \textbf{Weight} & \textbf{(\%)} & \textbf{Weight} & \textbf{(\%)} \\ 
\midrule
Imaging Quality & 0.70 & \textcolor{lowsatred}{15.28\%} & 0.75 & \textcolor{lowsatred}{17.95\%} \\
Aesthetic Quality & 0.76 & \textcolor{lowsatred}{16.55\%} & 1.05 & \textcolor{lowsatred}{25.06\%} \\
Subject Consistency & 0.51 & \textcolor{lowsatred}{11.02\%} & 1.00 & \textcolor{lowsatred}{23.88\%} \\
Background Consistency & 1.30 & \textcolor{lowsatred}{28.28\%} & 0.81 & \textcolor{lowsatred}{19.44\%} \\
Motion Smoothness & 0.84 & \textcolor{lowsatred}{18.34\%} & $-1.20$ & \textcolor{lowsatred}{0.00\%} \\
Temporal Flickering & $-1.00$ & \textcolor{lowsatred}{0.00\%} & 0.57 & \textcolor{lowsatred}{13.68\%} \\
Overall Consistency & 0.48 & \textcolor{lowsatred}{10.55\%} & 0.00 & \textcolor{lowsatred}{0.00\%} \\
\bottomrule
\end{tabular}}
\label{tab:vqa_metrics_analysis}
\end{table}


\subsection{Text-to-Video Retrieval}

\noindent\textbf{Implementation Details.} 
Our experimental setting evaluates the utility of the SynTVA dataset for the TVR task. We assess generalisation to real-world scenarios using the standard MSRVTT 1K-A test set \cite{MSRVTT1k} as our evaluation benchmark. All experiments employ the X-Pool \cite{gorti2022x_tvr} model, initialized with CLIP ViT-B/32 \cite{radford2021CLIP} pre-trained weights.
We train the X-Pool \cite{gorti2022x_tvr} model on various training subsets constructed from SynTVA. The construction of these subsets considers the following factors: \textbf{Semantic Alignment Quality.} `High-Quality' and `Low-Quality' subsets are created based on human annotation scores for each semantic alignment dimension. \textbf{Data Quantity.} For each quality type (High and Low), training subsets are formed in nine different sizes of video-text pairs. Category-based selection is applied to ensure the data balance. Furthermore, we use a `Random' baseline by randomly selecting videos for each of these nine quantities, irrespective of quality scores. To evaluate TVR performance, we report Recall@1 (R@1), as this metric provides a strict and clear measurement of retrieval effectiveness.

\noindent\textbf{High Semantic Alignment Positively Correlates with TVR Utility.} Figure~\ref{fig:TrainingQuantity} demonstrates a clear positive correlation between the assessed semantic alignment quality of synthetic videos and their resulting utility in TVR tasks. The strength of this positive impact is particularly pronounced for the \textit{Object \& Scene} and \textit{Action} alignment dimensions. Models trained on high-quality subsets for these aspects achieve substantial R@1 gains over those trained on low-quality subsets, reflected in performance gaps of 0.8 and 0.6, respectively. Therefore, video-text pairs from these two critical dimensions are denoted as HQ-SynTVA.

\noindent\textbf{High-Quality Synthetic Videos Enhance TVR Performance.} Figure \ref{fig:TrainingQuantity} confirms that high-quality synthetic videos serve as an effective data augmentation method for enhancing TVR performance. Even with a small quantity of 20 samples, models trained on high-quality synthetic data consistently outperform the pre-trained baseline across all alignment dimensions. This performance advantage generally scales with increasing data quantity for most dimensions. An exception is \textit{Prompt Fidelity}, which shows flatter gains, likely due to its consistently high human annotation scores (as shown in Figure \ref{fig:rader_chart}). Crucially, when considered in conjunction with our Auto-evaluator framework, this finding suggests a potential direction for leveraging appearance-based VQA metrics (e.g., from VBench) to efficiently generate and select high-utility synthetic videos, thereby improving TVR model performance.






\section{Conclusion}

In this paper, we introduce SynTVA: a benchmark dataset with 800 diverse queries, 2,400 video-text pairs and detailed human annotations across four semantic alignment dimensions for TVR. Our evaluation framework links standard VQA metrics to these alignment dimensions and features an Auto-Evaluator that predicts human-perceived alignment from VQA metrics. Experiments show synthetic videos with high \textit{Object \& Scene} and \textit{Action} alignment significantly boost TVR performance. Thus, our outcome demonstrates the value of high-quality synthetic data for augmenting TVR models, effectively bridging the gap between perceptual quality and task-specific utility.


\newpage
\balance
\bibliographystyle{ACM-Reference-Format}
\bibliography{reference}


\end{document}